\documentclass[dvipsnames]{article}

    \usepackage[final]{neurips_2024}

\usepackage{amsmath}
\usepackage{amssymb}
\usepackage{mathtools}
\usepackage{amsthm}
\usepackage{multirow}
\usepackage{makecell}
\usepackage{caption}
\usepackage{subcaption}
\usepackage{wrapfig}

\usepackage{graphicx}
\usepackage{pifont}

\usepackage[utf8]{inputenc} 
\usepackage[T1]{fontenc}    
\usepackage{url}            
\usepackage{booktabs}       
\usepackage{amsfonts}       
\usepackage{nicefrac}       
\usepackage{microtype}      

\usepackage{color, soul}
\usepackage[most,breakable]{tcolorbox}
\usepackage{xcolor}         

\usepackage[labelfont=bf]{caption}
\usepackage{enumitem}
\usepackage{sectsty}

\usepackage[colorlinks, linkcolor=RoyalBlue, anchorcolor=BrickRed, citecolor=RoyalBlue, urlcolor=RoyalBlue]{hyperref}
\usepackage{amsthm}
\newtheorem*{densinglaw}{Densing Law}
\newtheorem*{definition}{(Relative) Capability Density}
\newtheorem{corollary}{Corollary}

\newtcolorbox{mytheorem}{
  colback=gray!5, 
  colframe=gray!80, 
  boxrule=0.5pt, 
  arc=4pt, 
  left=4pt, 
  right=4pt, 
  top=4pt, 
  bottom=4pt, 
}

\title{Densing Law of LLMs}

\author{%
Chaojun Xiao\textsuperscript{\rm 1}, 
Jie Cai\textsuperscript{\rm 2},
Weilin Zhao\textsuperscript{\rm 1},
Guoyang Zeng\textsuperscript{\rm 2},
Biyuan Lin\textsuperscript{\rm 2},
Jie Zhou\textsuperscript{\rm 2},
Zhi Zheng\textsuperscript{\rm 2} \\
\textbf{
Xu Han\textsuperscript{\rm 1},
Zhiyuan Liu\textsuperscript{\rm 1}, Maosong Sun\textsuperscript{\rm 1}} \\
\textsuperscript{\rm 1}Tsinghua University~ \textsuperscript{\rm 2}ModelBest Inc. \\
\texttt{xiaocj20@mails.tsinghua.edu.cn} \\
\texttt{\{han-xu,liuzy,sms\}@tsinghua.edu.cn}
}

\begin{document}

\maketitle

\begin{center}
\section*{Highlights}
\end{center}

\vspace{-1em}
We introduce the concept of ``capability density'' to evaluate the training quality of large language models (LLMs) and describe the trend of LLMs that considers both effectiveness and efficiency.

\begin{mytheorem}
\begin{definition}
For a given LLM $\mathcal{M}$, its capability density is defined as the ratio of its \textbf{effective parameter size} to its actual parameter size, where the effective parameter size is the minimum number of parameters required for the reference model to achieve performance equivalent to $\mathcal{M}$.
\end{definition}
\end{mytheorem}

We reveal an empirical law for the capability density of \textit{\textbf{open-source base LLMs}} released since 2023.

\begin{mytheorem}
\begin{densinglaw}
The maximum capability density of LLMs exhibits an exponential growth trend over time.
\begin{equation*}
    \text{ln}(\rho_{\text{max}}) = A t + B
\end{equation*}
Here, $\rho_{\text{max}}$ is the maximum capability density of LLMs at time $t$.
\end{densinglaw}
\end{mytheorem}

Figure~\ref{fig:density} presents the capability density of popular LLMs, measured by their performance on $5$ widely-used benchmarks. 
A trend is fitted between maximum capability density and release date, revealing that $A \approx 0.007$ with $R^2 \approx 0.93$. This indicates \textbf{the maximum capability density of LLMs doubles approximately every $\textbf{3.3}$ months}\footnote{The capability density growth rate is affected by specific evaluation benchmarks and reference models.}. That means, around three months, it is possible to achieve performance comparable to current state-of-the-art LLMs using a model with half the parameter size.

\begin{figure}[htp]
\vspace{-1.0em}
    \centering
    \includegraphics[width=1.0\linewidth]{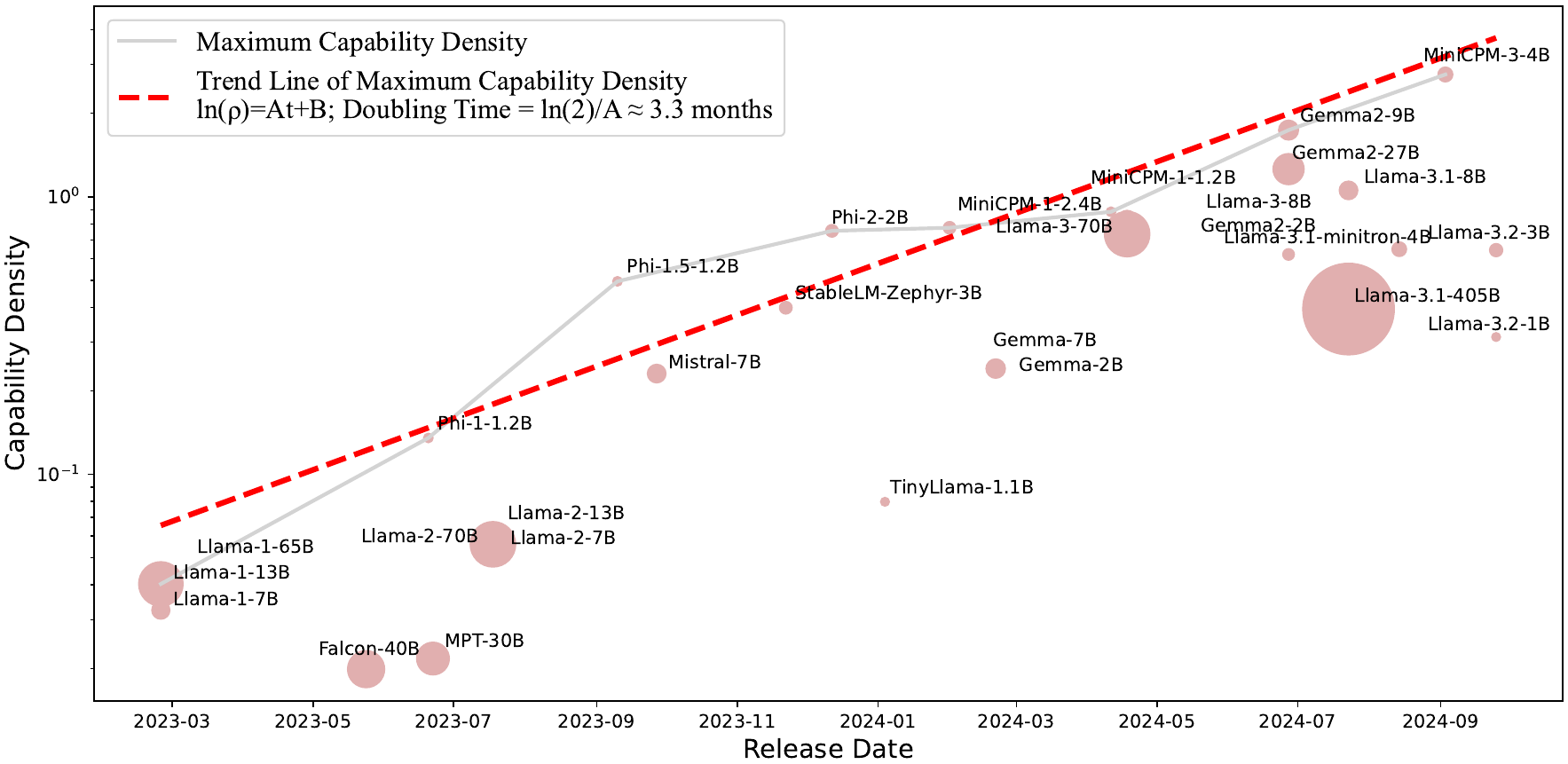}
    \caption{The estimated capability density of open-source base LLMs.}
    \label{fig:density}
\end{figure}


\newpage
\begin{abstract}
Large Language Models (LLMs) have emerged as a milestone in artificial intelligence, and their performance can improve as the model size increases.
However, this scaling brings great challenges to training and inference efficiency, particularly for deploying LLMs in resource-constrained environments, and the scaling trend is becoming increasingly unsustainable.
This paper introduces the concept of ``\textit{capability density}'' as a new metric to evaluate the quality of the LLMs across different scales and describes the trend‌ of LLMs in terms of both effectiveness and efficiency. 
To calculate the capability density of a given target LLM, we first introduce a set of reference models and develop a Scaling Law to predict the downstream performance of these reference models based on their parameter sizes. We then define the \textit{effective parameter size} of the target LLM as the parameter size required by a reference model to achieve equivalent performance, and formalize the capability density as the ratio of the effective parameter size to the actual parameter size of the target LLM.
Capability density provides a unified framework for assessing both model effectiveness and efficiency. Our further analysis of recent open-source base LLMs reveals an empirical law (Densing Law) that the capability density of LLMs grows exponentially over time. More specifically, using some widely used benchmarks for evaluation, the capability density of LLMs doubles approximately every three months. 
The law provides new perspectives to guide future LLM development, emphasizing the importance of improving capability density to achieve optimal results with minimal computational overhead.
\end{abstract}
\vspace{-1.5em}

\section{Introduction}
In recent years, large language models (LLMs) have garnered significant attention in the field of artificial intelligence, demonstrating remarkable improvements across various tasks~\citep{foundation-model,PTMs-survey-qiu,PTMs-survey,llama,gpt4}. The Scaling Law for LLMs further reveals that model performance continues to improve as model parameters and training data increase~\citep{scaling-law,scaling-law-multimodal,chinchilla}. This discovery has led to the development of LLMs with hundreds of billions of parameters, such as GPT-3 175B~\citep{gpt3}, PaLM 540B~\citep{palm}, and Llama-3.1-405B~\citep{llama3.1}, which have demonstrated exceptional capabilities in a wider range of applications.

Besides, with the advancement of LLMs, enhancing inference efficiency has become increasingly urgent:
1)~As LLMs are deployed in an expanding array of scenarios, inference costs have surpassed training costs, becoming the main bottleneck in practical applications~\citep{DBLP:conf/icml/SardanaPDF24,DBLP:journals/corr/abs-2404-02852,GPT-4o-mini}.
2)~There is a growing need to deploy LLMs on resource-constrained end devices like smartphones, serving as personal assistants, which requires models to be more efficient and compact~\citep{apple-intelligence,powerinfer2,minicpm}.
3)~The inference Scaling Law indicates that allowing LLMs to generate more tokens for "thinking" during the inference stage is crucial for improving performance in complex reasoning tasks~\citep{DBLP:journals/corr/abs-2407-21787,GPT-o1,DBLP:journals/corr/abs-2408-03314}, further increasing the demand for efficient inference.
To address these challenges, many efforts have been devoted to developing efficient LLMs with only billions of parameters to reduce inference overhead, such as OpenAI's GPT-4o-mini~\citep{GPT-4o-mini} and Apple's apple intelligence~\citep{apple-intelligence}.

Given these two seemingly contradictory paths – scaling up LLMs for effectiveness versus scaling down LLMs for efficiency – natural questions arise: 
\textit{Can we quantitatively evaluate the quality of LLMs with different scales? Is there a law that reflects the efficiency trend in LLMs, like the Scaling Law does for parameter and data scales?}

To this end, we introduce the concept of capability density, which serves as a metric for evaluating and comparing the training quality of LLMs on various scales.
Accurately measuring all aspects of an LLM's capabilities, or its level of intelligence, is quite challenging. In this article, we design a method to assess the relative capability density\footnote{For ease of explanation, in this work, we use ``density'' to refer to ``(relative) capability density''.}.
Specifically, we use a reference model and then estimate its scaling function between the performance on downstream tasks and parameter sizes. Based on the scaling function, for any given model, we calculate its effective parameter size – the number of parameters the reference model would need to achieve equivalent performance. The density of an LLM relative to the reference model is then defined as the ratio of its effective parameter size to its actual parameter size.
By introducing the concept of model density, we aim to more accurately measure model quality and enable comparisons between models of different scales. This evaluation method has the potential to provide new insights into the future direction of LLM development, helping researchers find the optimal balance between effectiveness and efficiency.

\subsection{Key Findings}
After defining LLM density, we analyze $29$ widely-used \textit{open-source pre-trained base models} from recent years. Our key finding for model density is:

\begin{mytheorem}
\begin{densinglaw}
The maximum capability density of LLMs exhibits an exponential growth trend over time.
\begin{equation*}
    \text{ln}(\rho_{\text{max}}) = A \cdot t + B
\end{equation*}
Here, $\rho_{\text{max}}$ is the maximum capability density of LLMs at time $t$.
\end{densinglaw}
\end{mytheorem}

Based on our evaluation on $5$ widely-used benchmarks, MMLU~\citep{mmlu}, BBH~\citep{bbh}, MATH~\citep{math-dataset}, HumanEval~\citep{humaneval}, and MBPP~\citep{mbpp}, $A \approx 0.007$, which means the maximum density of LLMs doubles approximately every three months. For example, MiniCPM-1-2.4B released on February 1st, 2024, can achieve comparable or even superior performance with Mistral-7B released on September 27th, 2023. We can use an LLM with only $35$\% parameters to obtain roughly equivalent performance after $4$ months.
It is worth noting that using different evaluation benchmarks may result in slight variations in the estimation and growth rate of model density. We encourage the community to develop more comprehensive evaluation benchmarks for LLMs to ensure more accurate measurements of density.

Based on the conclusion that the density of LLMs is continuously increasing in an exponential trend, we can further deduce the following implications:

\begin{mytheorem}
\begin{corollary}
\textbf{Inference Costs Decrease Exponentially}: The inference costs are going down exponentially for LLMs with equivalent downstream performance.
\end{corollary}
\end{mytheorem}
Densing Law indicates that the ratio of effective parameter size to the real parameter size doubles approximately every three months. Intuitively speaking, in three months, we can achieve performance comparable to the current state-of-the-art model using a model with only half the number of parameters. Thus, the inference costs are going down exponentially for equivalent downstream performance. We find that from January 2023 to the present, the inference cost of GPT-3.5-level models has decreased by 266.7 times.

\begin{mytheorem}
\begin{corollary}
\textbf{Densing Law $\times$ Moore's Law}: The effective parameter size of LLMs that can run on chips of the same area increases exponentially.
\end{corollary}
\end{mytheorem}
Moore's Law~\citep{moorelaw} states that the number of circuits integrated on a chip of the same area increases exponentially. This implies an exponential increase in computing power. Densing Law indicates that the density of LLMs doubles every $3.3$ months. Combining these two factors, we can conclude that the effective parameter size of LLMs that can be run on a chip of the same price increases faster than both LLMs' density and computation power of chips.

\begin{mytheorem}
\begin{corollary}
\textbf{Density Growth Accelerated after ChatGPT's Release}: With the release of ChatGPT, the growth rate of LLM density increased by $50\%$.
\end{corollary}
\end{mytheorem}
We compare the increasing trends in LLMs' density before and after the release of ChatGPT. The results show that following the release of the ChatGPT model, the growth rate of maximum density has noticeably accelerated. Specifically, after the release of ChatGPT, the growth rate of LLM density increased by $50\%$.

\begin{mytheorem}
\begin{corollary}
\textbf{Efficient Compression $\neq$ Density Improvement}: Existing pruning and distillation methods usually cannot lead to efficient LLMs with higher density.
\end{corollary}
\end{mytheorem}
To enhance model inference efficiency, many researchers have devoted efforts to a series of model compression algorithms, such as pruning and distillation~\citep{ma2023llm,sunsimple,yang2024survey,xu2024survey}. These algorithms are often believed to improve the performance of the resulting compressed models. However, by comparing some models with their compressed counterparts, we can observe that the widely used pruning and distillation methods usually result in smaller models with lower density than the original models. We encourage the community to further explore more effective model compression algorithms, with a greater emphasis on improving the density of smaller models.

\begin{mytheorem}
\begin{corollary}
\textbf{Towards Density-Optimal Training - Green Scaling Law}: The development of LLMs should shift from being performance-centric to being density-centric.
\end{corollary}
\end{mytheorem}
Density is a metric that reflects the trade-off between effectiveness and efficiency. Therefore, blindly increasing model parameters to pursue performance improvements can lead to lower model density, resulting in unnecessary energy consumption. For example, while Llama-3.1-405B~\citep{llama3.1} achieves state-of-the-art performance among open-source models, it requires computational resources that are hundreds of times greater than other models. Consequently, model developers need to shift their focus from merely optimizing performance to optimizing density. This approach aims to achieve the best results with minimal computational costs, thereby realizing a more sustainable and environmentally friendly Scaling Law.

\vspace{0.5em}
In this work, we propose a new evaluation metric, capability density, for LLMs, which can offer a new, unified perspective on the two current trends – enhancing effectiveness and increasing efficiency. 
Based on our proposed metric, we evaluate $29$ open-source models and find an empirical experience law, named Densing Law: the density of LLMs exhibits an exponentially increasing trend. Based on this empirical relationship, we discuss several deductions and provide observational evidence. 
Through this novel evaluation perspective, we hope to provide valuable insights and guidance for the future development of LLMs.

\section{Density for Large Language Models}
In this section, we formally define the density for LLMs, which is calculated as the ratio of the effective parameter size to the actual parameter size.  In the following sections, we will first describe the overall framework and formal definition of LLM density. Then we introduce how to utilize the Scaling Law to estimate the effective parameter size.

\subsection{Overall Framework and Definition}
The core of LLM density lies in the effective parameter size, which refers to the number of parameters required for a reference model to achieve the same performance as a given model.
To achieve this, we need to fit a function that relates the parameter sizes of the reference model to its performance. Specifically, for a given model $\mathcal{M}$ with $N_\mathcal{M}$ parameters, assume its performance score on the downstream tasks is $S_\mathcal{M}$. This score can be calculated using various metrics depending on the downstream task, such as accuracy, F1 score, etc. To compute the effective parameter size, we train a series of reference models with varying scales of parameters and training data. Based on these models, we fit a function between the parameter size and performance: $S = f(N)$, where $S$ denotes the downstream performance, and $N$ represents the parameter sizes of the reference model. Then we can calculate the effective parameter size as $\hat{N}(S) = f^{-1}(S)$ and the density for $\mathcal{M}$ is defined as:
\begin{equation}
    \rho(\mathcal{M}) = \frac{\hat{N}(S_\mathcal{M})}{N_\mathcal{M}} = \frac{f^{-1}(S_\mathcal{M})}{N_{\mathcal{M}}}.
\end{equation}

It is important to note that Scaling Laws are typically used to fit the relationship between language modeling loss and parameter sizes~\citep{scaling-law}, and it is non-trivial to predict downstream task performance directly. Inspired by Llama-3~\citep{llama3.1}, we adopt a two-step estimation approach: (1)~\textbf{Loss Estimation}: In the first step, we use a series of reference models to fit the relationship between the parameter size and language modeling loss on the test set, expressed as $\mathcal{L} = f_1(N)$. (2)~\textbf{Performance Estimation}: Due to the presence of emergent abilities~\citep{emergent-ability}, it is challenging to accurately estimate the relationship between parameter sizes and performance using reference models with limited training computes. Therefore, we incorporate open-source models to compute their loss and performance on the test set and fit the relationship $s = f_2(\mathcal{L})$. This two-step estimation process allows us to derive $s = f_2(f_1(N))$. In the following sections, we will provide a detailed description of the fitting processes for $f_1(\cdot)$ and $f_2(\cdot)$.

\subsection{Loss Estimation}
To predict the performance of downstream tasks, the first step involves fitting a function between the parameter size and language model loss using the Scaling Law widely adopted for LLM pre-training. Previous Scaling Laws primarily focus on language modeling loss on the whole sequences, which reflects the model's ability to estimate the probability of a given corpus. However, instances in the downstream tasks usually encompass both input instructions and output answers and we are primarily concerned with the probability of the output answers. Therefore, in this work, we focus on fitting the conditional loss $\mathcal{L} = -\text{log}(P(\text{answer} \mid \text{instruction}))$. Concretely, we estimate a power-law function between the conditional loss $\mathcal{L}$, and parameter size $N$, as well as the number of training tokens $D$:
\begin{equation}
\label{eq:loss}
    \mathcal{L} = a N^{-\alpha} + b D^{-\beta},
\end{equation}
where $a$, $\alpha$, $b$, and $\beta$ are parameters need to be fitted. 

In previous research on Scaling Laws~\citep{scaling-law}, the loss typically needs to be specified on a validation corpus, and the average loss is calculated over all tokens in this corpus. In this work, our goal is to fit the model's performance on downstream tasks, which require models to output the answers based on the input instructions. Therefore, we directly calculate the conditional loss on downstream tasks, meaning the loss incurred by the model when generating answers given the task inputs.
(1)~For multiple-choice problems, calculating the loss solely based on the content of the correct option can lead to inaccurate estimates, as it ignores the content of incorrect options. Besides, if we only calculate the loss on the final option labels, the loss for single token is also unstable. Therefore, we concatenate the problem and its multiple options as inputs, and the output is the analysis for the input problem as well as the final answer label.
(2)~For most complex problems, such as mathematical questions, we often require the model to generate a sequence of reasoning steps before providing the final answer. For these tasks, when calculating the loss, we include both the reasoning steps and the correct answer as the output to compute the model's loss. It is important to note that most datasets do not provide reasoning steps for each instance. For both two types of tasks, we use GPT-4o~\citep{gpt4} to generate reasoning steps for all test instances.
These approaches allow us to better estimate the model's performance by considering the specific requirements and formats of different tasks.


\subsection{Performance Estimation}
In the second step, we need to predict downstream task performance based on the loss on test sets. In the loss estimation step, the Scaling Law models trained with limited training computes usually cannot achieve meaningful scores on downstream tasks, with most Scaling Law models performing only at the level of random guessing. Thus, it is impossible to predict the downstream performance with only these models. 
To address this issue, we incorporate well-trained open-source models for function fitting and calculate their loss and performance on the test set.
Considering that the performance for most downstream tasks is bounded, we use a sigmoid function for fitting. The sigmoid function naturally maps all input values to the range of 0 to 1. Additionally, when the loss is particularly large, the model's performance should approximate that of random guessing, and when the loss is particularly small, the model's performance should approach the upper bound. This characteristic aligns with the properties of the sigmoid function, which is very flat at both extremes of the curve. Specifically, we estimate the downstream performance with the following function:
\begin{equation}
\label{eq:performance}
    S = \frac{c}{1 + e^{-\gamma (\mathcal{L} -l)}} + d,
\end{equation}
where $c$, $\gamma$, $l$, and $d$ are parameters need to be estimated. 



\begin{table}[b]
    \centering
    \small
    \caption{The detailed hyper-parameters of small models trained for loss estimation. }
    \begin{tabular}{rrrrrrrrr}
    \toprule
    Name & \# Para & BS & $n_{layer}$ & $d$ & $d_{ffn}$ & $d_{head}$ & $n_{head}$ & $n_{kv}$ \\ \midrule
    0.005B & 5,247,232   &  32 & 8  &   256 &  640  & 64 & 4  & 1 \\
    0.03B  & 31,470,080  &  32 & 12 &   512 & 1,280 & 64 & 8  & 2 \\
    0.1B   & 106,196,736 &  64 & 18 &   768 & 1,920 & 64 & 12 & 3 \\
    0.2B   & 245,416,960 & 128 & 24 & 1,024 & 2,560 & 64 & 16 & 2 \\
    0.4B   & 476,852,480 & 256 & 30 & 1,280 & 3,200 & 64 & 20 & 2 \\
    0.8B   & 828,225,024 & 512 & 36 & 1,536 & 3,840 & 64 & 24 & 3 \\
    \bottomrule
    \end{tabular}
    \label{tab:scaling-model}
    \vspace{-1em}
\end{table}

\subsection{Density}
After fitting Equation~\ref{eq:loss} and \ref{eq:performance}, given the performance $S_\mathcal{M}$ of a model $\mathcal{M}$, we can infer the effective parameter size by utilizing the inverse functions of these equations. It is important to note that in Equation~\ref{eq:loss}, the loss $\mathcal{L}$ is a bivariate function of both the parameter count $N$ and the training data size $D$. Therefore, when calculating the effective parameter size, it is necessary to specify a particular training data size $D$. Here, to calculate the effective parameter size, we defaultly use $D = D_0 = 1T$ tokens. Then the effective parameter size can be explained as the parameter size the reference model trained with $D_0$ tokens needs to achieve equivalent performance. Concretely, we can compute the effective parameter size as:
\begin{equation}
     \hat{\mathcal{L}}(S_\mathcal{M}) = l - \frac{1}{\gamma}ln\left(\frac{c}{S_{\mathcal{M}} - d} - 1\right);\quad
     \hat{N}(S_\mathcal{M}) = \left(\frac{\hat{\mathcal{L}}(S_\mathcal{M}) - bD_0^{-\beta}}{a}\right)^{-\frac{1}{\alpha}}.
\end{equation}
Now, we have established the relationship between the downstream performance and effective parameter size. The density of the given model $\mathcal{M}$ is $\rho(\mathcal{M}) = \frac{\hat{N}(S_\mathcal{M})}{N_\mathcal{M}}$. Intuitively, if one model can achieve better performance with the same scale of parameters, then the model's density is higher. Therefore, in the future, considering the limited computation resources of deployment devices, we should devote great effort to improving the model's density instead of merely increasing the model parameter scales for better performance.

\begin{figure*}
    \centering

    \begin{subfigure}[b]{0.97\textwidth}
    \centering
    \includegraphics[width=\textwidth]{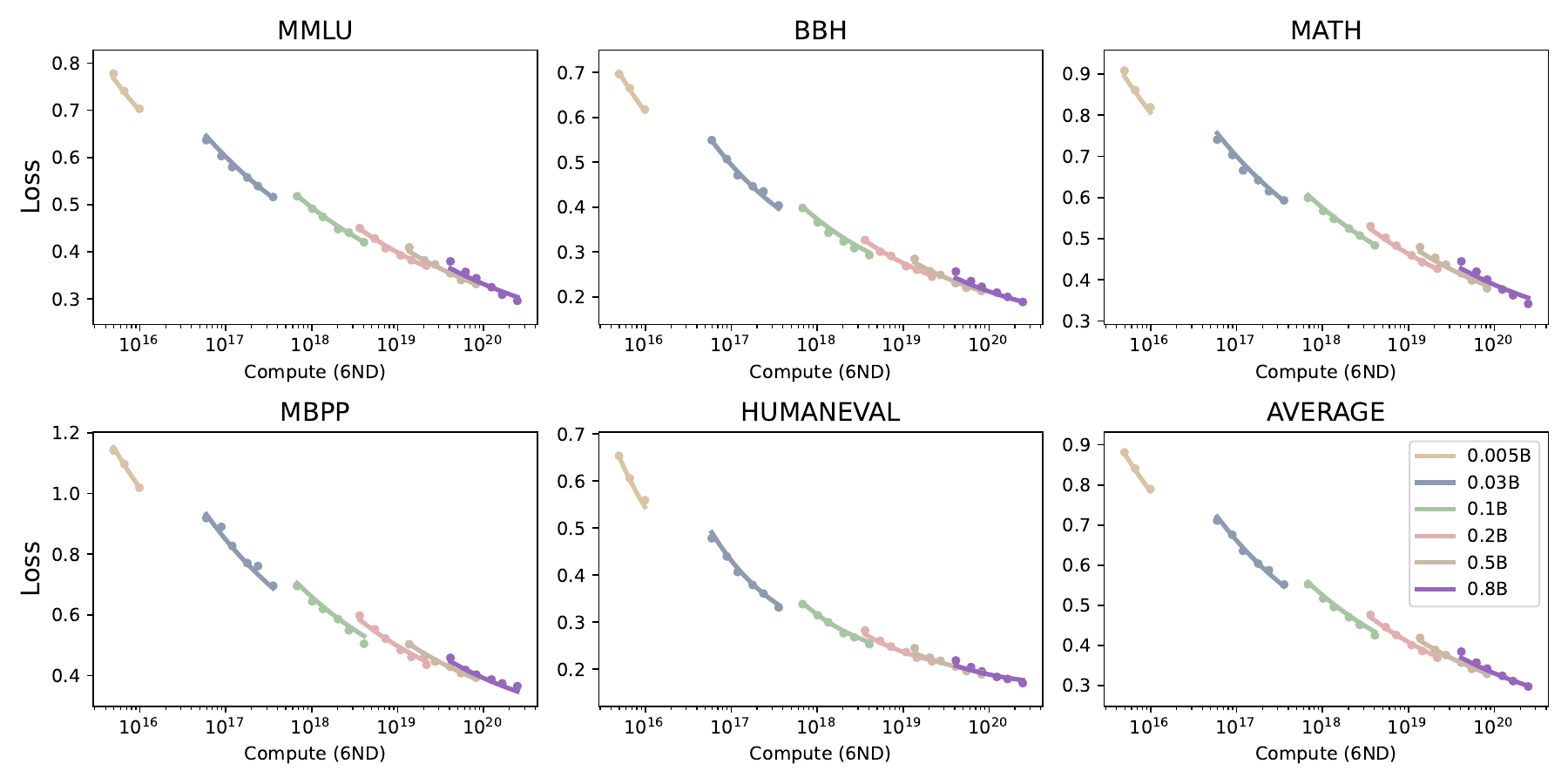} 
    \caption{Loss Estimation}
    \end{subfigure}
    \begin{subfigure}[b]{0.97\textwidth}
    \centering
    \includegraphics[width=\textwidth]{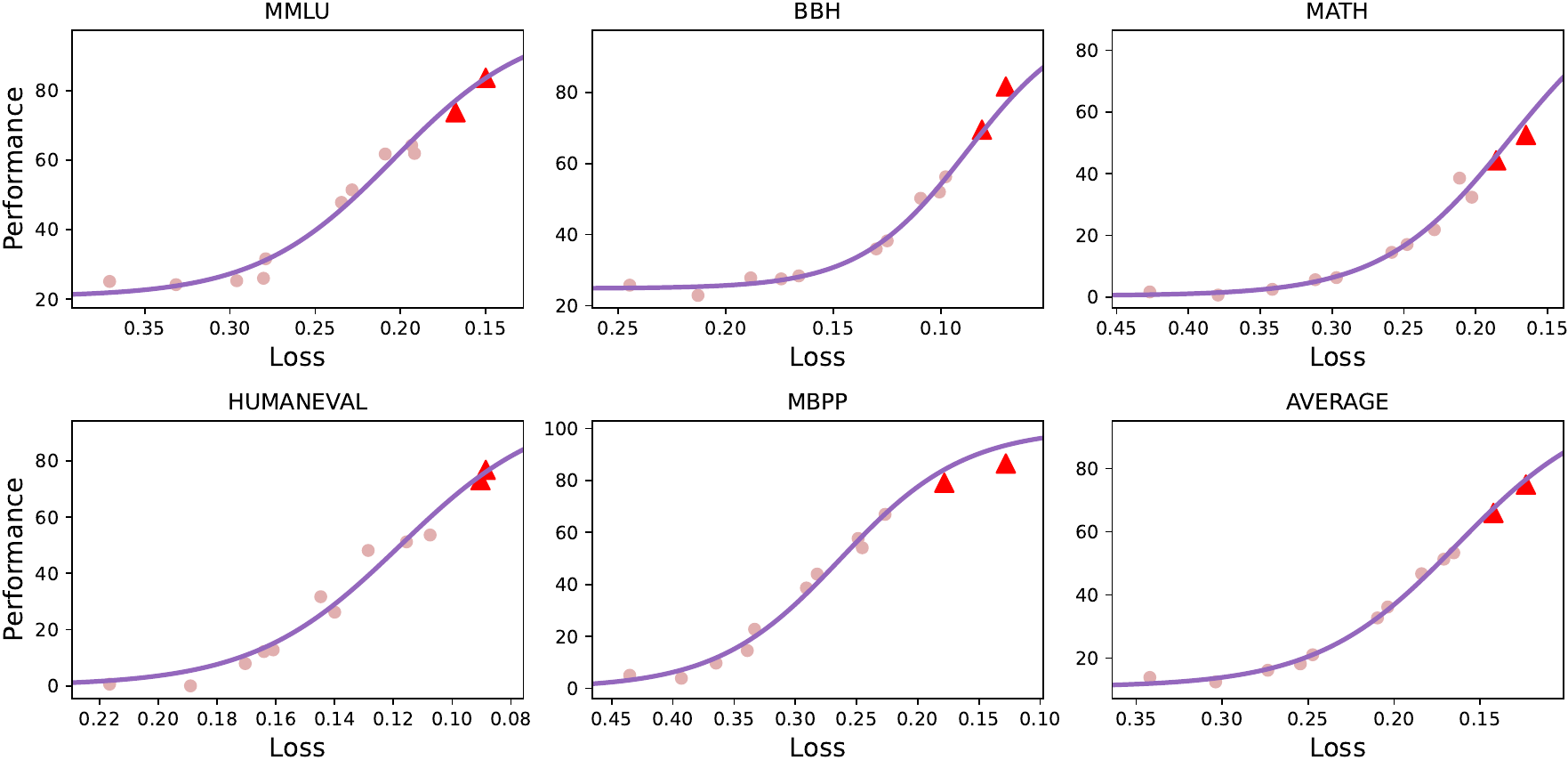} 
    \caption{Performance Estimation}
    \end{subfigure}
    \caption{The results for loss estimation and performance estimation. Here, the lines are fitted curves. X-axis in (a) refers to the pre-training compute, which is approximated by $\text{Compute}=6ND$. Triangles in (b) are larger models for prediction.}
    \label{fig:estimation-results}
    \vspace{-0.5em}
\end{figure*}


%


\section{Density Evolution}

\subsection{Evaluation Settings}
\textbf{Dataset}
In this work, we adopt the following widely-used datasets for evaluation: MMLU~\citep{mmlu} for English knowledge-intensive tasks, BBH~\citep{bbh} for challenging logic reasoning tasks, 
MATH~\citep{math-dataset} for mathematical reasoning tasks, and HumanEval~\citep{humaneval}, MBPP~\citep{mbpp} for coding tasks. We apply the open-source tools~\citep{opencompass,evalplus} for evaluation. Here, we evaluate all models in a few-shot in-context learning manner and these models are required to generate the final answer label based on the given demonstrations and inputs of test instances. Following widely-used settings, MMLU, BBH, MATH, HumanEval, and MBPP are evaluated under the $5$-shot, $3$-shot, $4$-shot, $0$-shot, and $3$-shot settings, respectively. Besides, for BBH, MATH, and MBPP, we adopt the chain-of-thought prompting technique~\citep{cot}.

\textbf{Loss Estimation Models}
In the loss estimation step, we need to run a series of models with different scales of parameters and training data. These models will be used as the reference models for further density computation. In this work, we adopt the training corpus of MiniCPM-3-4B~\citep{minicpm}, a widely-used edge-size model, to train the small models. As for the model architecture, we use grouped query attention~\citep{gqa}, gated feedforward layers with SiLU as the activation function. We train the models using Warmup-Stable-Decay learning rate scheduler. To estimate the scaling curve, we train the models with $\{10, 15, 20, 30, 40, 60\}\times N$ tokens, where $N$ refers to the parameter size. We list the hyper-parameters for small scaling models in Table~\ref{tab:scaling-model}.

\textbf{Performance Estimation Models}
In the performance estimation step, we introduce additional well-trained models to fit the loss-performance curve. Specifically, we use a series well-trained MiniCPM-3 models and their intermediate training checkpoints. Their parameter scales range from $0.5$ billion to tens of billion. These models use the same vocabulary as our scaling models with different parameter sizes and training datasets. 

\textbf{Evaluated Models}
Furthermore, to illustrate the change in density over time, we select widely used LLMs for evaluation since the release of Llama-1~\citep{llama}, as most open-source models released before Llama-1 cannot achieve meaningful performance on our selected datasets. Specifically, we evaluate the density of the following models: Llama series of models~\citep{llama,llama2,llama3.1}, Falcon~\citep{falcon}, MPT~\citep{MPT}, Phi series of models~\citep{phi,phi1.5,phi3}, Mistral~\citep{mistral}, StableLM~\citep{stablelm}, TinyLlama~\citep{tinyllama}, and MiniCPM series of models~\citep{minicpm}.
We prioritize using the results reported in each model's technical reports for density calculations. Besides, we only evaluate the density of base pre-trained models without instruction tuning as the instruct-tuning datasets may contain human-annotated data similar to our selected test data leading to inaccurate density estimation.
Notably, many pre-trained models also introduce supervised finetuning datasets in the pre-training phase, leading to the test set contamination issue~\citep{wei2023skywork,dominguez2024training}. Thus, the inaccurate density estimation remains to be solved, which we leave for future work.

Notably, we only evaluate the density of pre-trained base models without further supervised fine-tuning and preference learning, due to the following reasons:
(1)~Pre-trained base models serve as the foundation for model performance. Considering the impact of further alignments, such as the quality of human annotations and the choice of alignment algorithms, introduces excessive confounding factors unrelated to the capabilities of the base model itself. 
(2)~The Scaling Law for the performance of LLMs with alignment remains an open question that requires further exploration.
Nowadays, there are numerous methods to improve the performance during inference time, such as retrieval-augmented generation~\citep{rag}, and thinking more for inference Scaling Law~\citep{GPT-o1}. Here, we only consider the basic prompting technique for base LLM evaluation, as this technique cannot consistently improve the performance of this base model. And we leave the density calculation for different inference FLOPs for future work, which may lead to inference Densing Law.

\subsection{Loss and Performance Estimation Results}
We present the estimation results of the two-step process in Figure~\ref{fig:estimation-results}. From the results, we can observe that the two-step estimation process can effectively fit the performance of different-sized models on three downstream tasks. With the decrease in the loss on the test instances, the performance significantly improves as a sigmoidal curve, and the loss has a power-law relationship with the number of parameters and training tokens.

To evaluate the effectiveness of our estimated method, we use models with parameters of less than $4$ billion to fit the loss-performance curve and preserve larger models for prediction. Triangles in Figure~\ref{fig:estimation-results}(b) are two models with tens of billions of parameters. From the results, we can observe that we effectively predict the downstream performance based on the loss values.

\subsection{Densing Law}
After fitting the loss scaling curve and the performance scaling curve, we further measured the density of widely used open-source models since the release of Llama-1~\citep{llama}. We present the density of each model along with their release dates in Figure~\ref{fig:density}. From the figure, we can observe that:
(1)~The density of LLMs has rapidly increased over time. Notably, the density of Llama-1, released in February 2023, is below $0.1$, whereas more recently released models like Gemma-2-9B and MiniCPM-3-4B have densities reach $3$. This increase in density is largely attributed to the growth in the scale of pre-training data and improvements in the quality of that data. For example, Llama-1 is pre-trained on 1.4 trillion tokens, whereas Llama-3 utilizes 15 trillion tokens with careful data cleaning.
(2)~Better performance does not always lead to better density. Llama-3.1-405B is currently one of the state-of-the-art open-source models due to its large-scale parameters. However, it is not the model with the highest density. This is because constrained by computational resources and the scale of pre-training data, we usually cannot fully optimize the training settings for extremely large models, making them sub-optimal in terms of cost-effectiveness.


To further illustrate the growth trend of the LLMs' density, we perform a linear fit on the envelope line in Figure~\ref{fig:density}. Specifically, we assume that the logarithmic value of the maximum density increases linearly over time. Formally, we fit the following linear function:
\begin{equation}
    \text{ln}(\rho_{\text{max}}) = A \cdot t + B,
\end{equation}
where $t$ is the time interval (unit: days) since the release date of Llama-1, $\rho$ is the maximum density value at time $t$, and $A, B$ are the parameters to be fitted. Through the fitting process, we obtained $A \approx 0.0073$, which implies that the density of the large model doubles approximately every $\frac{\text{ln}(2)}{A} \approx 95 $ days. Here, the $R^2$ for the linear regression function is $0.912$. 

The growth trend in model density reveals an important pattern in the development of current LLMs. While Scaling Laws indicate that model performance improves with an increase in parameter size, the parameter scale growth is constrained by the limited computation resources available in deployment scenarios and the demand for fast response. As a result, large models are not simply evolving towards larger parameter sizes. Instead, developers of LLMs are striving for higher cost-effectiveness, aiming to achieve optimal performance with minimal inference costs. This discovery aligns with the principles discovered by Moore's Law in the development of integrated circuit chips~\citep{moorelaw}, which emphasize increasing transistor density on a limited chip area. Therefore, we name our discovery on the growth trend in model density as Densing Law.

\subsection{Corollaries of Densing Law}

Based on Densing Law and our evaluation results, in this section, we discuss several corollaries and hope our discovery can promote the development of LLMs.

\textbf{Inference Costs Decrease Exponentially} 
The density of LLMs shows an exponential growth trend, doubling approximately every three months. Here, density is defined as the ratio of the effective parameter size to the actual parameter size. This implies that in three months, we can achieve performance comparable to current models using only half the actual parameter size. Consequently, under the condition of achieving the same performance, the actual parameter size of LLMs will also decrease exponentially. This reduction in actual parameter count translates to decreased computational costs during inference. Therefore, the exponential increase in LLMs' density will directly result in an exponential decrease in inference costs for models achieving the same level of performance.

\begin{wrapfigure}{r}{0.5\textwidth}
  \centering
    \includegraphics[width=\linewidth]{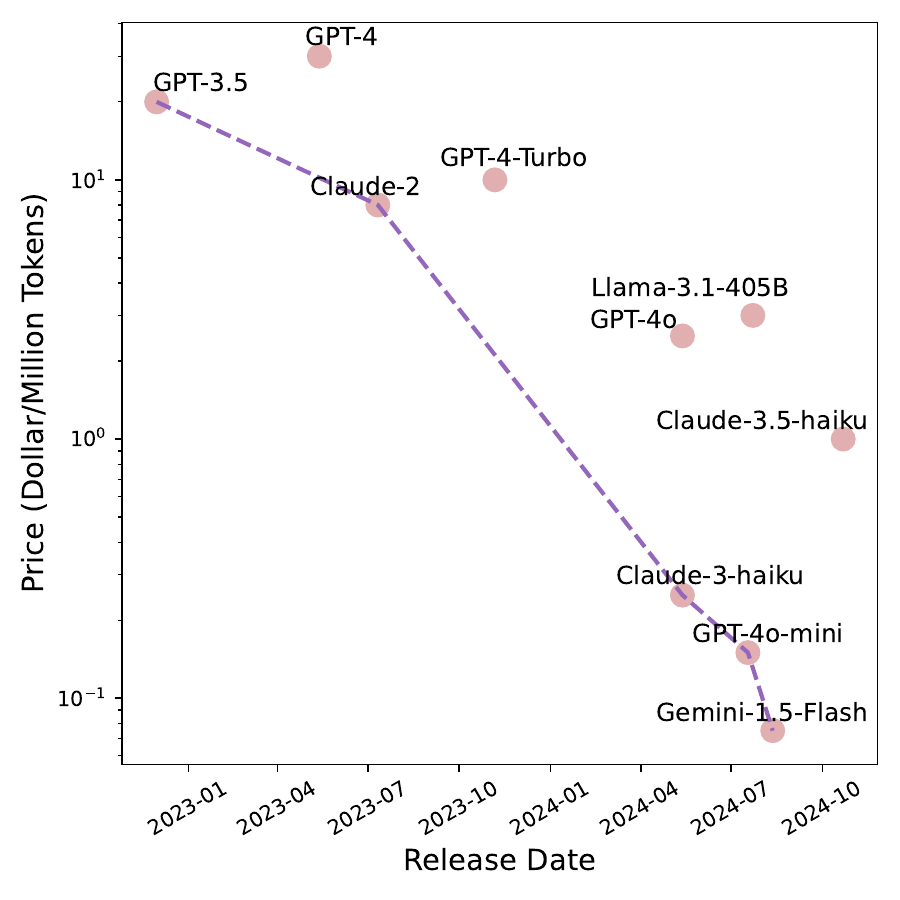}
    \caption{Prices of LLMs that can outperform GPT-3.5. The line connects the cheapest models.}
    \label{fig:price}
\end{wrapfigure}

To better illustrate the decreasing trend in inference costs for LLMs, we present the API pricing of LLMs that have achieved superior performance to GPT-3.5 since its release in Figure~\ref{fig:price}. From the figure, we can observe that the prices of LLMs exhibit an exponential decline. Specifically, in December 2022, GPT-3.5 cost \$20 for one million tokens, whereas by August 2024, Gemini-1.5-Flash costs only \$0.075 for the same number of tokens, a reduction of $266.7$ times. Roughly speaking, the inference costs for LLMs halve approximately every 2.6 months. The exponentially decreasing trend of LLM API pricing is also observed in~\cite{infcost}.

In addition, we can observe that the rate of decline in inference costs is faster than the growth rate of LLMs' density. This is because inference costs depend not only on the actual parameter size but also heavily on the inference infrastructure. In recent years, inference systems for LLMs have garnered significant attention from researchers, including optimizations in memory access speed for self-attention layers~\citep{vllm,flashattention,flashattention2} and sparse computation optimizations for feed-forward networks~\citep{powerinfer,liu2023deja}. These advancements have greatly contributed to the reduction in inference costs for LLMs.

\textbf{Densing Law Meets Moore's Law}
Densing Law describes the exponential trend of increasing model density over time, focusing on improvements at the algorithmic level of LLMs. On the other hand, Moore's Law, which states that computing power increases exponentially, highlights advancements in hardware technology~\citep{moorelaw}. The combination of these two principles suggests a rapidly approaching future where high-quality LLMs can run efficiently on consumer-grade devices, such as smartphones and PCs, with low power consumption. This convergence of algorithmic efficiency and hardware capability is paving the way for more accessible and widespread use of advanced AI technologies in everyday devices.

Specifically, recent observations~\citep{epoch2023trendsinmachinelearninghardware} found that the computation power of chips with the same price doubles approximately every $2.1$ years.
Densing Law indicates that the ratio between the effective parameter size and the actual parameter size doubles every three months. Therefore, given a fixed chip price, the effective parameter size of the largest LLM that can run on it grows exponentially. This growth rate is the product of the growth rate of model density and the growth rate of transistor density on the chip. Based on current estimates, this implies that the maximum effective parameter size approximately doubles every 88 days. This rapid growth highlights the combined impact of advancements in both algorithmic efficiency and hardware technology, suggesting a future where increasingly powerful models can be deployed on existing hardware much more quickly than previously anticipated.



\begin{figure*}
    \centering
    \includegraphics[width=\linewidth]{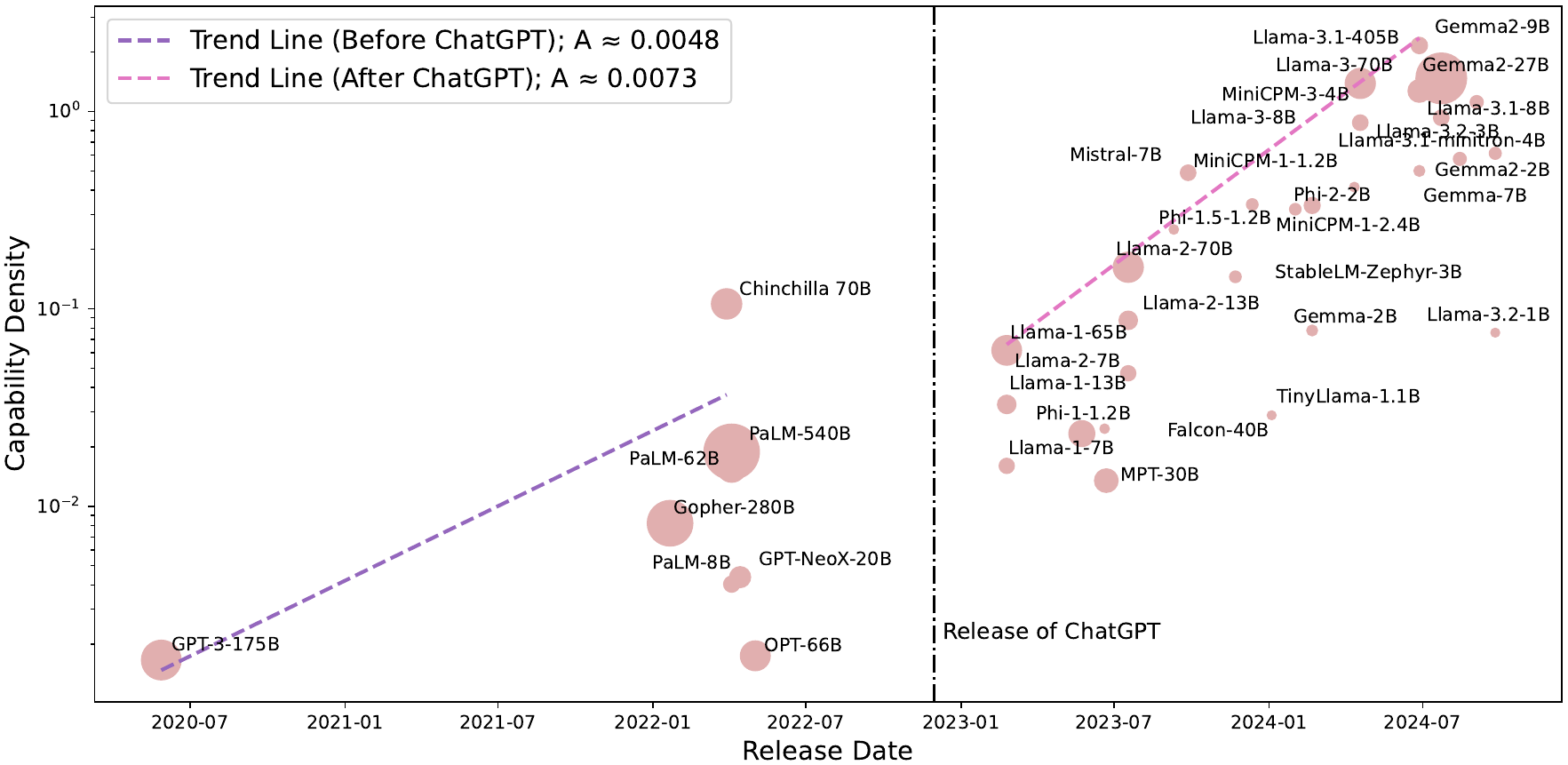}
    \caption{Density evaluated using MMLU. Two trend lines represent the growth of LLMs' density before and after the release of ChatGPT.}
    \label{fig:growth-change}
    \vspace{-1em}
\end{figure*}

\textbf{Density Growth Accelerated after ChatGPT's Release}
In 2022, ChatGPT achieved great performance improvements across various tasks and its zero-shot generalization ability spurred significant efforts from both industry and academia to advance the development of LLMs. To illustrate the change in the trend of model density growth before and after the release of ChatGPT, we evaluate the densities of typical LLMs since the release of GPT-3. We use the MMLU benchmark to capture the changes in density. The results are presented in Figure~\ref{fig:growth-change}.

From the figure, we can observe that the rate of increase in model density significantly accelerated following the release of ChatGPT. Before ChatGPT, the slope of the trend line was approximately $A \approx 0.0048$, whereas after its release, it increased to $A \approx 0.0073$, indicating a $50\%$ faster growth rate in model density. Several factors contribute to this accelerated growth:
(1)~Increased investment: The success of ChatGPT highlighted the potential of LLMs, leading to a significant increase in investment directed towards LLM development.
(2)~More high-quality open-source models: The rise in high-quality open-source models has lowered the barriers to research and development in LLMs. After ChatGPT's release, there was a notable increase in high-quality small LLMs with only billions of parameters, whose accessibility allows many researchers to conduct LLM research using relatively small GPU clusters. Therefore, we encourage the community to open-source their cutting-edge algorithms and models, which can significantly contribute to density improvement.

\begin{wrapfigure}{r}{0.5\textwidth}
  \centering
    \includegraphics[width=\linewidth]{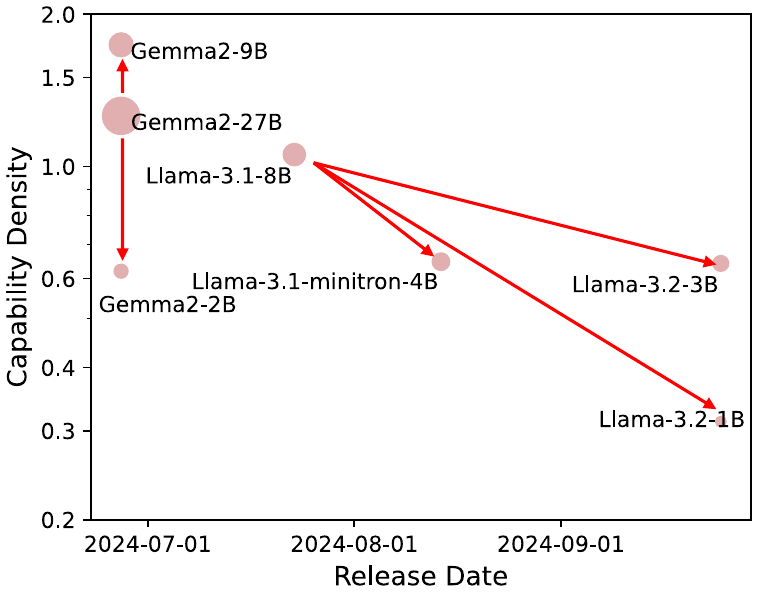}
    \caption{Comparison between compressed models and their larger counterparts.}
    \label{fig:pruning}
\end{wrapfigure}

\textbf{Efficient Compression $\neq$ Density Improvement}
LLMs are often constrained by high inference costs, making it challenging to run them on consumer devices. To address this issue, many developers employ pruning and distillation techniques to compress LLMs. In Figure~\ref{fig:pruning}, we also present the densities of several compressed models. For instance, Llama-3.2-3B/1B and Llama-3.1-minitron-4B~\citep{minitron} are derived from pruning and distilling Llama-3.1-8B~\citep{llama3.1}, while Gemma-2-9B/2B is distilled from Gemma-2-27B~\citep{gemma2}.

The results show that only the Gemma-2-9B model has a higher density than the original model, whereas all other compressed models have lower densities compared to their original counterparts. Intuitively, pruning involves removing unimportant neurons from LLMs, which suggests that these neurons might store less knowledge than other neurons. This would imply that compressed models should intuitively achieve higher density. However, the results are quite the opposite. This discrepancy might be due to the insufficient training of smaller models during the compression process, preventing them from reaching optimal density. Therefore, we encourage the community to address this challenge by ensuring that compressed models are adequately trained during future efforts.

\textbf{Towards Density-Optimal Training - Green Scaling Law}
Since the release of GPT-3~\citep{gpt3} and the introduction of the Scaling Law~\citep{scaling-law}, many researchers focus on training language models with extremely large parameter sizes to continuously enhance model performance. Guided by this trend, PaLM-540B~\citep{palm} and Gopher-280B~\citep{gopher} achieve great improvements on various natural language processing tasks. Given the constraints of pre-training computational resources, maximizing the use of pre-training clusters to develop training compute-optimal LLMs has become a key focus~\citep{chinchilla}. Furthermore, inference compute costs have surpassed training compute costs as a major concern, leading to a shift towards pre-training smaller models using increasingly large-scale training data~\citep{minicpm,apple-intelligence}.

In light of the discovery of the Densing Law, we now encourage a shift towards density-optimal LLM pre-training. With the continuous efforts in LLM development worldwide, model density is rapidly increasing, resulting in shorter lifecycles for each model. Simply increasing the scale of pre-training corpora for LLMs can lead to longer development cycles and higher training costs. However, shortly after a model is released, it is expected that a new model with comparable performance and lower inference costs will be available in three months. In this context, LLM developers must consider the growth trend of model density and adopt more efficient and generalized training techniques to enhance model density. This approach helps avoid excessive cost investments and the losses associated with short profit recovery cycles.

\vspace{-1em}







\section{Discussion}

\textbf{Accurate Capability Measurement}
Capability density reflects the abilities of an LLM per unit of parameters. However, with current technology, we cannot accurately assess the absolute capability level of LLMs, meaning that quantifying intelligence remains a great challenge. Therefore, in this work, we design a method to measure the relative density value of LLMs. Besides, we use widely-used benchmarks to evaluate the performance of LLMs. However, the limited number of benchmarks and potential data contamination issues introduce bias in performance evaluation. Thus, advancing accurate measurement of LLMs' capabilities or intelligence levels in the future will enable better calculation of their density.

\textbf{Connection between Densing Law and Scaling Law}
The Scaling Law of LLMs reveals the relationship between an LLM's performance and its parameter and data sizes, reflecting the intrinsic characteristics of complex systems composed of vast numbers of neurons. The Densing Law further highlights the trend in the development of LLMs' efficiency and effectiveness over time, marking a technological advancement trend as humanity pursues high-level AI models. Formally, under conditions of sufficient training data, the Scaling Law explains the relationship between model loss and parameter size as: $\mathcal{L} = AN^{-\alpha}$, which is appropriate for the training of all Transformer-based models. Furthermore, the Densing Law indicates that developers of LLMs can increase $\alpha$ through continuous improvements in data, algorithms, and architecture, thereby reducing the model loss for a given parameter size.


\textbf{Period of Validity of Densing Law}
Densing Law reveals the rapid development of LLM algorithms. In this paragraph, we discuss the question: \textit{how long this exponential growth in model density will continue?}. We believe that the rapid increase in model density is driven by significant investments in personnel and resources. The improvement of general intelligence capabilities in LLMs can bring substantial benefits to various industries, further encouraging investment in model research and development.
Given the great potential of LLMs, we believe that Densing Law will remain effective for a considerable period. However, it is essential to continually update the evaluation datasets used to evaluate model density, as LLMs will soon achieve satisfactory performance on existing datasets.
In the event of achieving artificial general intelligence, LLMs themselves may be capable of conducting scientific research autonomously, exploring new pathways to further increase density. At that point, the growth in LLM density could accelerate even more, driven by the models' ability to innovate and optimize their own development processes.


\section{Limitations and Future Directions}
In this section, we discuss the limitations and future directions of our proposed method to evaluate the capability density of LLMs.

\textbf{Fair and Comprehensive Evaluation}
The capability density measurement of LLMs relies on existing benchmarks to evaluate model performance. Therefore, the benchmark quality greatly impacts the density measurement results. In this work, we use those benchmarks widely adopted by researchers to evaluate various LLMs. However, several challenges remain: 
(1)~Comprehensive evaluation: With the development of LLMs, the capabilities of LLMs significantly expand, such as the ability to handle complex reasoning tasks~\citep{GPT-o1}. Consequently, the capability density measurement needs to be continually updated by incorporating more comprehensive evaluation datasets that reflect evolving capabilities.
(2)~Fair evaluation: With the increasing scale of pre-training data and the construction of synthetic data, some LLMs are overoptimized towards benchmarks, leading to inflated scores. To address this, we plan to use newly constructed datasets to evaluate model performance, thereby mitigating the overfitting risk and ensuring accurate density estimation.

\textbf{Multi-modal Density}
In this work, we focus on measuring the capability density of language models. However, measuring the density and trends in large multimodal models is also crucial as multimodal applications increase. In the future, designing reasonable density evaluation methods for multimodal models will be an important research direction.

\textbf{Inference Densing Law}
Recent research has highlighted that more inference computational costs allow LLMs to engage in deeper reasoning, effectively enhancing their performance on complex tasks~\citep{GPT-o1}. In this work, we use the parameter size as the basis to evaluate model capability density.
However, as the importance of chain-of-thought reasoning continues to grow, density evaluation should shift towards being based on inference FLOPs. Specifically, capability density could be formalized as the ratio of effective inference FLOPs to actual inference FLOPs. In this way, we hope that LLMs achieve optimal results with the minimum number of reasoning steps.


\section{Conclusion}

To illustrate the recent trend towards efficient LLMs and to quantitatively measure the training quality of LLMs, this paper introduces a method for evaluating the capability density of LLMs. By measuring the capability density of open-source base LLMs released since 2023, we show an empirical law: the capability density of LLMs increases exponentially over time. The evaluation results on some widely-used LLM benchmarks indicate that the density of LLMs doubles every three months. This implies that, within three months, a model with only half the parameters can achieve performance comparable to the current state-of-the-art models. This finding highlights the rapid development and increasing efficiency of LLMs.
We discuss several corollaries based on the law, and hope that the law and its corollaries will encourage the LLM community to continue enhancing model capability density and achieving optimal performance with minimal computational costs.

\bibliographystyle{citation}
\bibliography{citation}

\end{document}